\documentclass[letterpaper]{article} 
\usepackage{aaai2026}  
\usepackage{times}  
\usepackage{helvet}  
\usepackage{courier}  
\usepackage[hyphens]{url}  
\usepackage{graphicx} 
\usepackage{natbib}  
\usepackage{caption} 
\usepackage{algorithm}
\usepackage{algorithmic}

\usepackage{newfloat}
\usepackage{listings}
\DeclareCaptionStyle{ruled}{labelfont=normalfont,labelsep=colon,strut=off} 
\floatstyle{ruled}
\newfloat{listing}{tb}{lst}{}
\floatname{listing}{Listing}
\usepackage{algorithm}
\usepackage{algorithmic}
\usepackage{amsmath}
\usepackage{amssymb}
\usepackage{amsfonts}       
\usepackage{booktabs}
\usepackage{multirow}
\DeclareMathAlphabet{\mathbbold}{U}{bbold}{m}{n}

\newcommand{\ie}{\textit{i.e.}}

\newcommand{\eg}{\textit{e.g.}}
\newcommand{\name}{RMMSS}

\title{\name: Towards Advanced Robust Multi-Modal Semantic Segmentation with \\ Hybrid Prototype Distillation and Feature Selection}
\author{
    Jiaqi Tan$^{1}$,
    $^\dagger$Xu Zheng$^{2,3}$,
    Yang Liu$^{1,}$\thanks{Corresponding author $^\dagger$ Project Lead}
}
\affiliations{
    \textsuperscript{\rm 1}School of Digital Media and Design Art, Beijing University of Posts and Telecommunications\\

    \textsuperscript{\rm 2} HKUST(GZ)
    \textsuperscript{\rm 3} INSAIT, Sofia University “St. Kliment Ohridski”\\
}

\usepackage{bibentry}

\begin{document}

\maketitle
\begin{abstract}
Multi-modal semantic segmentation (MMSS) faces significant challenges in real-world applications due to incomplete, degraded, or missing sensor data. While current MMSS methods typically use self-distillation with modality dropout to improve robustness, they largely overlook inter-modal correlations and thus suffer significant performance degradation when no modalities are missing. 
To this end, we present \textbf{\name}, a two-stage framework designed to progressively enhance model robustness under missing-modality conditions, while maintaining strong performance in full-modality scenarios. 
It comprises two key components: the \textbf{H}ybrid \textbf{P}rototype \textbf{D}istillation \textbf{M}odule (HPDM) and the \textbf{F}eature \textbf{S}election \textbf{M}odule (FSM). In the first stage, we pre-train the teacher model with full-modality data and then introduce HPDM to do cross-modal knowledge distillation for obtaining a highly robust model.
In the second stage, we freeze both the pre-trained full-modality teacher model and the robust model and propose a trainable FSM that extracts optimal representations from both the feature and logits layers of the models via feature score calculation. This process learns a final student model that maintains strong robustness while achieving high performance under full-modality conditions. Our experiments on three datasets demonstrate that our method improves missing-modality performance by 2.80\%, 3.89\%, and 0.89\%, respectively, compared to the state-of-the-art, while causing almost no drop in full-modality performance (only -0.1\% mIoU). Meanwhile, different backbones (AnySeg amd CMNeXt) are utilized to validate the generalizability of our framework.

\end{abstract}


\section{Introduction}

Multi-modal semantic segmentation (MMSS), as a fundamental task in computer vision~\cite{long2015fully,liu2016exploration} effectively overcomes the perception limitations of single modality by collaboratively integrating various data streams coming from the sensors such as depth information~\cite{wang2021brief,zhou2020rgb}, LiDAR point clouds~\cite{camuffo2022recent} and event streams~\cite{alonso2019ev}. 
In recent years, researchers have made significant progress in multi-modal fusion methods under ideal conditions, demonstrating superior segmentation performance~\cite{zhang2023cmx,zhang2023DELIVERing,reza2024mmsformer}. However, real-world scenarios introduce 0significant challenges, such as incomplete, degraded, or missing sensor data, all of which can lead to data loss or distortion, thereby reducing the performance of MMSS.

The observed performance degradation primarily arises from the model's robustness in integrating multi-modal information, which manifests in two critical ways: \textcircled{1} Most training paradigms assume complete multi-modal inputs, causing models to overfit to idealized conditions and perform poorly under real-world challenges such as sensor failure cases. For instance, CMNeXt~\cite{zhang2023DELIVERing} exhibits a substantial mIoU drop from 66.30 to 48.77 when transitioning from ideal RDEL to DEL settings. \textcircled{2} Variations in the informativeness of different modalities give rise to unimodal bias\cite{huang2022modality,kleinman2023critical,peng2022balanced} a tendency for models to over-rely on dominant, easily learnable modalities such as RGB\cite{dancette2021beyond,gat2020removing,han2021greedy}, undermining the balanced use of available sensory inputs.

To address the robustness issue caused by missing modalities, existing multi-modal vision methods primarily focus on modality masking~\cite{shin2024complementary} and self-distillation~\cite{zheng2024learning,hou2024sdstrack}. However, these approaches operate mainly at the feature level and fail to leverage the inherent properties of visual distillation. Furthermore, they often suffer from a limitation of knowledge distillation: \textbf{\textit{a noticeable performance drop under full-modality conditions}}. 
In multi-modal fusion, cross-modal distillation has demonstrated effectiveness in reducing modality gaps such as in text-image tasks~\cite{gu2025breaking}. Nevertheless, prior research in MMSS has mostly explored intra-modal distillation (e.g., RGB-to-RGB). In contrast, we observe that hybrid distillation strategies—such as RGB-to-Depth and Depth-to-LiDAR are more effective in capturing the complementary relationships across modalities, thereby significantly enhancing model robustness.


To bridge these gaps, we introduce a two-stage framework, namely \name, to progressively enhance model robustness under missing-modality conditions, while maintaining strong multi-modal performance. Specifically, in stage 1, we focus on maximizing the model’s resilience to incomplete modality input. To achieve this, we adopt a self-distillation strategy based on modality masking, and introduce a novel Hybrid Prototype Distillation Module (HPDM). HPDM encodes modality-specific features from the backbone into prototype representations, and performs hybrid distillation between student and teacher prototypes by randomly pairing different modality combinations. This encourages the model to learn robust representations regardless of modality availability. At this stage, our model \textbf{has become the SoTA} in terms of robustness (+2.73\%mIoU compare to AnySeg~\cite{zheng2024learning}).
In stage 2, the goal is to retain the robustness gained from stage 1 while further optimizing for complete-modality performance. The model trained in stage 1 acts as a robust teacher, and is complemented by a second teacher trained with full-modality input. Together, these two teachers jointly supervise the student model. To effectively combine their knowledge, we introduce a Feature Selection Module (FSM), which selectively integrates output features from both teachers. This fused representation captures the complementary strengths of robustness and rich modality information, and serves as the supervisory signal for the student model.

As shown in Table~\ref{tab:datasets}, extensive experimental results on three public benchmarks show that 
our model boosts mIoU by 2.80\%, 3.89\%, and 0.89\% without increasing the number of model parameters, respectively, while sacrificing only 0.05–0.25 \% mIoU compare with the full-modality teacher.
In summary, our contributions are threefold:
(I) We propose a two-stage framework for robustness enhancement named \textbf{\name}, which can improve the robustness of any multi-modal semantic segmentation model.
(II) We are the first to introduce cross-modal distillation into MMSS and propose the \textbf{Hybrid Prototype Distillation Module (HPDM)}, which transforms features into prototype representations and leverages cross-modal distillation to assist the self-distillation framework in alleviating modality bias in multi-modal semantic segmentation tasks, thereby enhancing robustness.
(III) We propose the \textbf{Feature Selection Module (FSM)}, which effectively balances knowledge from multiple teachers while preserving full-modality performance and robustness under missing modalities.

\begin{figure*}[ht!]
\centering
\includegraphics[width=1.0\textwidth]{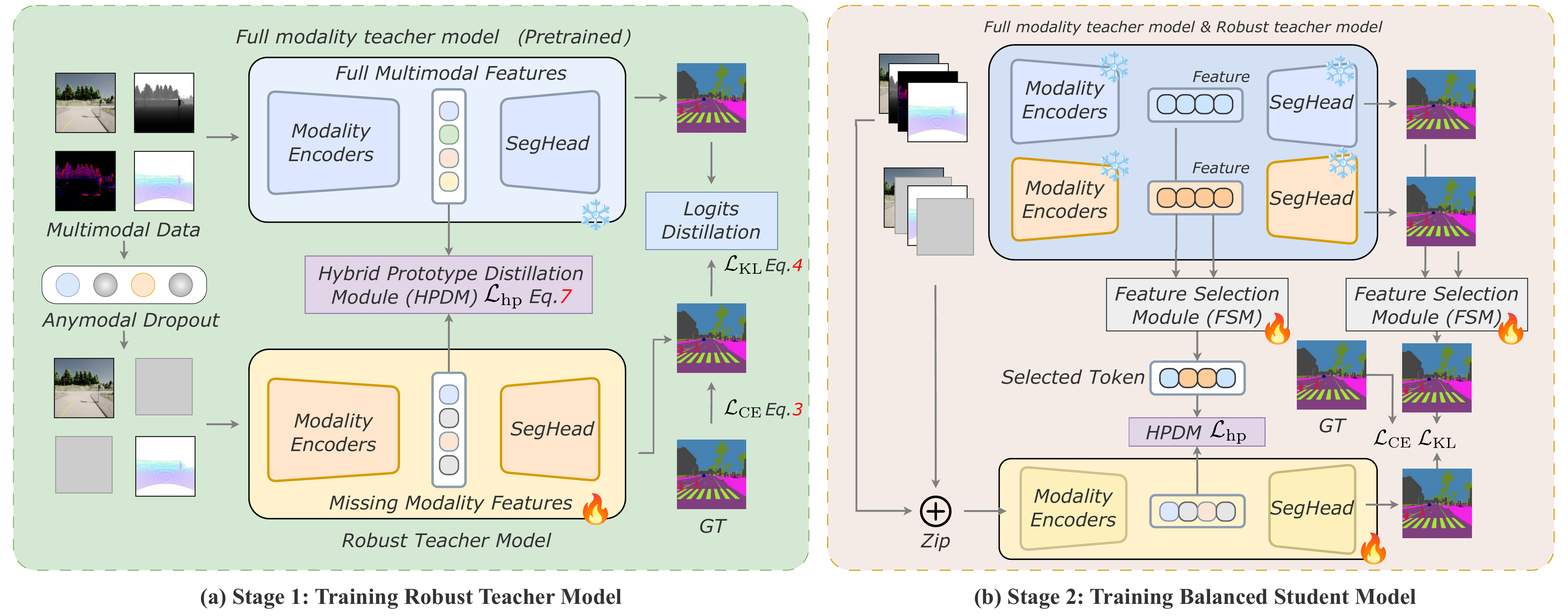}
\caption{Overview of the \name~framework.}
\label{fig:RobustSeg}
\end{figure*}

\section{Related Work}
\subsection{Robust Multi-Modal Semantic Segmentation}
Multi-Modal Semantic Segmentation, the task of assigning a class label to each pixel, has evolved from early single-modal architectures like Fully Convolutional Networks (FCNs)~\cite{liu2015parsenet,noh2015learning} and Feature Pyramid Networks (FPNs)~\cite{chen2014semantic,chen2018encoder,lin2017feature} to more recent Transformer-based models~\cite{strudel2021segmenter,xie2021segformer,zheng2021rethinking} that leverage multi-scale perception and contextual attention.  
Incorporating additional sensor data (e.g., Depth, LiDAR) has been shown to significantly improve segmentation performance~\cite{ gao2021we,wang2021brief,zhang2023cmx}, making multi-modal segmentation a key research direction. However, robustness remains a critical, yet underexplored, area for real-world applications.  
Existing multi-modal segmentation models can be broadly classified into three types~\cite{liao2025benchmarking}:  
1) \textbf{RGB-centric models} prioritize RGB as the primary input, using supplementary modalities for refinement (\eg, CMNeXt~\cite{zhang2023DELIVERing} with a Self-Query Hub). 
2) \textbf{Modality-agnostic models} treat all inputs equally, fusing multi-level features (\eg, AnySeg~\cite{zheng2024learning}).  
3) \textbf{Adaptive models} dynamically select the dominant modality based on feature similarity~\cite{zheng2024magic++,li2024stitchfusion}. In this work, we adopt the second approach at knowledge distillation stage, promoting equal treatment of all modalities to reduce unimodal bias and enable effective cross-modal distillation.
\subsection{Knowledge Distillation}

The knowledge distillation paradigm follows the teacher-student learning strategy, which transfers the knowledge acquired by a teacher network or an ensemble of networks to another network ~\cite{hinton2015distilling, ba2014deep}. Some research reformulates supervision signals to enhance knowledge transfer efficacy, including probability transfer~\cite{passalis2018learning}, relational transfer~\cite{park2019relational}, and correlation transfer~\cite{peng2019correlation}, while conducting domain-specific exploration of signal distributions in 
segmentation~\cite{wang2020intra,zhou2022rethinking}. Due to its excellent knowledge transfer capability, knowledge distillation has been applied in the forms of self-distillation and cross-modal distillation in recent studies. In self-distillation, perturbations such as modality masking and noisy inputs are introduced during student training, while a full-modality teacher model is used to guide the student toward correct learning~\cite{zhou2023unidistill,shin2024complementary,hou2024sdstrack}. This approach has shown promising results in enhancing robustness, particularly in adversarial image learning~\cite{zhao2024mitigating} and medical image segmentation tasks~\cite{wang2023prototype}. However, this kind of framework will encounter the problem of performance degradation in the full-modality. In cross-modal knowledge transfer, knowledge distillation is commonly employed for balancing and aligning representations across different modalities, such as Audio-Visual~\cite{chen2021distilling} and Text-Visual~\cite{gu2025breaking}. However, there is no cross-modal research in MMSS at present.

\section{Method}
\label{sec:methodology}



\subsection{Preliminary}
The initial self-distillation and modality masking framework typically consists of a teacher model pretrained on full-modality data and a student model trained with randomly dropped modalities. Modality masking prevents the model from overfitting to full-modality data, while the teacher model provides guidance during training, alleviating performance degradation. Based on the above equations, the self-distillation loss can be simply expressed as the sum of the the student-label loss and teacher-student loss. The specific cross-entropy \(\mathcal{L}_{\text{CE}}\) loss is formulated as follows:
\begin{equation}
    \mathcal{L}_{\text{CE}}=-\frac{1}{N}\sum_{n=1}^{N}\sum_{c=1}^{C}y_{n,c}\text{log}(\text{P}(x_{n,c})),
\end{equation}
and the \(\mathcal{L}_{\text{KL}}\) loss is formulated as:
\begin{equation}
    \mathcal{L}_{\text{KL}}= \frac{1}{N}\sum_{n=1}^{N}\sum_{c=1}^{C}\text{P}(l_s^{n,c})\text{log}\frac{\text{P}(l_s^{n,c})}{\text{P}(l_t^{n,c})},
\end{equation}
where \(N\) spans the minibatch dimension, \(C\) is the number of classes, \(\text{P}\) is the softmax function. For Cross-Entropy,  \(x\) and \(y\) are the logits corresponding to class \(c\) of the n-th minibatch from the student model and teacher model, respectively. For KL Divergence, \(l_s\) and \(l_t\) are modality features or final logits from the student model and teacher model. We construct the basic loss function for the student model, as shown in Eq.~\ref{eq:lossorigin}:
\begin{equation}
\mathcal{L}_{\text{origin}}=\mathcal{L}_{\text{CE}}+\lambda \mathcal{L}_{\text{KL}} ,
\label{eq:lossorigin}
\end{equation}
here, \(\lambda\) is a balancing parameter that adjusts the weight between Cross-Entropy loss and KL divergence loss.

\subsection{Overview}
Although the self-distillation framework is effective, it has limitations when applied to multi-modal semantic segmentation (MMSS) tasks that rely heavily on complementary information across modalities. 
In this paper, we introduce \name, a general framework designed to enhance model robustness. In this section, we first present the details of the two-stage \name~framework (Section \ref{sec:\name}). We then delve into the Hybrid Prototype Distillation Module (HPDM) (Section \ref{sec:HPDM}), where we explore how to integrate self-distillation, cross-modal distillation, and the specific characteristics of segmentation tasks to improve model robustness. Finally, in Section~\ref{sec:FSM}, we propose the Feature Selection Module (FSM), which facilitates balanced multi-teacher distillation in the second stage.

\subsection{\name~Framework}
\label{sec:\name}

Our \name~consists two stages: In the first stage, we train a model with strong robustness; in the second stage, a balanced model is trained under the joint supervision of a full-modality model and the robust model. These models maintain the same architecture during training. 
In the first stage, as shown in Figure~\ref{fig:RobustSeg} (a), to better balance information across modalities through cross-modal distillation, we select the encoded features from each modality produced by the backbone network as the targets for distillation. These features contain rich initial semantic information specific to each modality. However, directly applying cross-modal distillation on raw features often leads to catastrophic performance degradation due to their high complexity. Existing cross-modal distillation methods typically compress modality-specific features into compact representations before conducting information transfer. 

Motivated by prototypical adaptation in dense prediction tasks~\cite{wang2020intra,zhou2022rethinking}, we input modality features \(\{f_r^i,f_d^i, f_e^i, f_l^i, \ldots\}\), extracted from the multi-modal input \(\{x_r,x_d,x_e,x_l,\ldots\}\), into the HPDM for prototype extraction. Cross-modal prototype distillation is then performed using randomly paired modalities between the teacher and student models. Notably, as shown in Table~\ref{tab:2stage}, our model already achieves state-of-the-art robustness at this stage and outperforms larger models (e.g., MiT-B2) across various metrics, despite its smaller size (MiT-B0).


In the second stage, as shown in Figure~\ref{fig:RobustSeg} (b), to mitigate performance degradation under full-modality conditions, we use the robust model from stage one and a full-modality model as complementary teachers. Each excels at different types of information, so we introduce the Feature Selection Module to fuse informative features from both teachers. This enables joint supervision and helps train a balanced student model. We fuse the obtained information and train the final model by guiding the student network—using the HPDM and logits in the same first stage manner as before. Please refer to Appendix A for details on Data Processing.

\subsection{Hybrid Prototype Distillation Module}
\label{sec:HPDM}
Prototypical features are more compact than pixel features and mine semantic associations in data more efficiently~\cite{wang2023prototype,wang2020intra}. In cross-modal knowledge transfer scenarios, due to modality differences, cross-modal distillation requires a semantic-level alignment and fusion to make knowledge transfer more natural and effective. Motivated by this, we employ prototypical features to achieve enhanced effectiveness in cross-modal knowledge transfer. In our method, we solve for prototypical features by calculating the similarity between each pixel and its corresponding class prototype. Specifically, we first adjust the original labels \( l \in \text{H} \times \text{W}\) to the size of the feature map \(f_m^i \) by interpolating the nearest neighbor, resulting in the interpolated and aligned labels  \(l' \in h \times w\). For a feature from the \(i^{th}\) stage of the \(m^{th}\) modality with class \(C\). The prototyping operation \(p\) is formulated as follows:
\begin{equation}
    p=[p_0,p_1,\ldots,p_C],\quad p_c=\frac{\sum_j f_m^{i,j}\mathbbold{1}[l_j' = c]}{\sum_j \mathbbold{1}[l_j' = c]},
\end{equation}
where \(f_m^{i,j}\) is the i-th stage feature of the pixel \(j\) of \(f_m^i\), \(l_j'\) is the ground truth of pixel \(j\). \(\mathbbold{1}\), which denotes the indicator function, outputs 1 if the specified condition holds true, and 0 otherwise. The output is \(p \in [C\times d]\), \(d\) is the same as the four stage feature dimensions \([32,64,160,256]\).

\begin{figure}[t!]
\centering
\includegraphics[width=8cm]{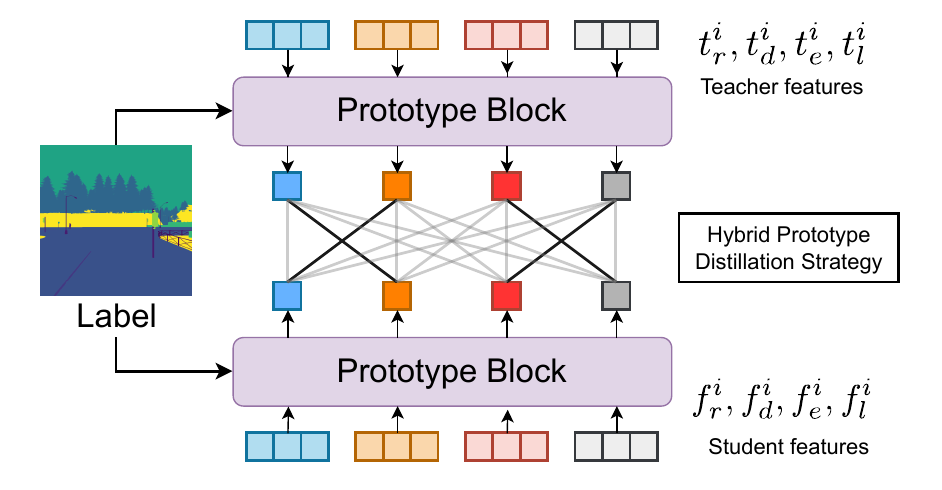}
\caption{The structure of HPDM. The features \(f\) from the teacher and student models are transformed into compact prototype features \(p\) via Eq.~\ref{eq:losshp}, after which a random matching strategy is applied to 
\(p\) for distillation.}
\label{fig:HPDM}
\end{figure}

Our \name~draws inspiration from cross-modal distillation~\cite{xia2023robust,zhou2023unidistill}. 
As shown in Tab ~\ref{fig:HPDM},to enhance interaction between modalities, we introduce a hybrid mechanism during prototype distillation. For each modality, we randomly permute the order of the features before computing prototypical representations. This enables each modality's prototypes to learn from others in a shuffled manner, enriching cross-modal knowledge transfer. Specifically, the hybrid prototype distillation loss is formulated as:
\begin{equation}
    \mathcal{L}_{\text{hp}}=\frac{1}{N}\sum_{n=1}^{N}\sum_{i=1}^{4}\sum_{m=1}^{M}\text{KL}(p^{n,i}_{\pi(m)},g^{n,i}_{m}),
    \label{eq:losshp}
\end{equation}
where \(N\) spans the batch size dimension, i refers to the feature stage, \(\pi(m)\) is function which returns a random number between 1 and \(M\) to shuffle the order of feature inputs (\eg, randomly altering the fixed feature sequence \([f_r^i, f_d^i, f_e^i, f_l^i]\) to \([f_l^i, f_r^i, f_e^i, f_d^i]\)). \(p^{n,i}_{\pi(m)}\) represents the result of applying the prototyping operation to the \(i^{th}\) stage student feature \(f_{\pi(m)}^i\) of the \(\pi(m)^{th}\) modality. \(g^{n,i}_m\) represents the result of applying the prototyping operation to the teacher feature \(t_{m}^i\) of the \(m^{th}\) modality correspondingly. Please refer to Appendix B.1 for details on HPDM dimension handling.

\subsection{Feature Selection Module}
\label{sec:FSM}
\begin{figure}[t!]
\centering
\includegraphics[width=7cm]{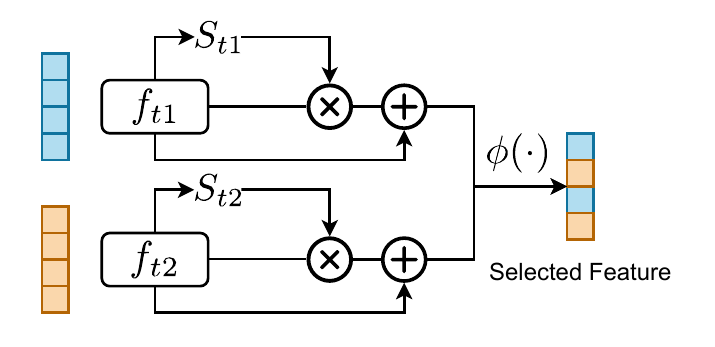}
\caption{The network design of the FSM Module. The input information \(f_{t1}\) and \(f_{t2}\) from different teachers will undergo score predictor and select the maximum value.}
\label{fig:FSM}
\end{figure}

As shown in~\ref{fig:FSM}, to fuse the complementary features from different teachers and fully leverage the distinct strengths of the robust teacher and the full-modality teacher, we introduce FSM for better knowledge selection. Given modality features zip \(f_t=[ f_{t1} , f_{t2}] \) from two different teacher models, each with a shape of \([H, W, C]\) and normalized to a unified scale, the FSM computes feature importance scores using a depthwise convolution layer followed by a sigmoid activation function. The computation is defined as follows:
\begin{equation}
\hat{f}_t = \text{DepthwiseConv}_{3 \times 3}(2,2)(f_t),
\end{equation}
\begin{equation}
S_t = \sigma\left(\text{Conv}_{1 \times 1}(2,1)(\hat{f}_t)\right),
\end{equation}
where \(\text{DepthwiseConv}_{3 \times 3}(\cdot)\) denotes a depthwise convolution operation that applies a \(3 \times 3\) kernel independently to each input channel (\ie, with groups equal to the number of channels \(2\)); 
\(\text{PointwiseConv}_{1 \times 1}(\cdot)\) refers to a \(1 \times 1\) convolution used to fuse information across channels, reducing the number of channels from \(2\) to \(1\), and \(\sigma(\cdot)\) denotes the sigmoid activation function. The Output can be formalized as:
\begin{equation}
f_t' = \phi\left(f_t , f_t + S_t \cdot \hat{f}_t\right)
\end{equation}
where  \(\phi(f,m)\) is a masking function that selects elements from \(f\) based on the position of the maximum value in \(m\).


\noindent \textbf{Overall Training Objectives:} Combining Eq.~\ref{eq:lossorigin} and \ref{eq:losshp}, we can obtain the  total loss function for two stages as follows:
\begin{equation}
    \mathcal{L} = \mathcal{L}_{\text{CE}} + \lambda\mathcal{L}_{\text{KL}} + \alpha \mathcal{L}_{\text{hp}},
\end{equation}
where \(\lambda\) and \(\alpha\) are hyperparameters that balance the contributions of loss terms. Two FSM modules respectively perform selection on multi-teacher prototype features and logits based on  \(\mathcal{L}_{\text{hp}}\) and \(\mathcal{L}_{\text{KL}}\). For more details on FSM gradient updates and data processing, please refer to Appendix B.2.

\section{Experiments}
\label{sec:experiments}
\subsection{Experimental Setup}
\label{sec:experimentalsetup}

\begin{table*}[t!]
\renewcommand{\tabcolsep}{4pt}
\resizebox{\linewidth}{!}{
\begin{tabular}{l|ccccccccccccccc|c}
\midrule
\multirow{2}{*}{Method} & \multicolumn{15}{c}{Anymodal Evaluation} & \multirow{2}{*}{Mean}\\ \cmidrule{2-16}
 & R & D & E & L & RD & RE & RL & DE & DL & EL & RDE & RDL & REL & DEL & RDEL & \\ \midrule
CMNeXt & 17.55 & 44.69 & 4.98 & 5.56 & 60.19 & 18.49 & 17.93 & 45.21 & 45.65 & 4.97 & 60.28 & 60.54 & 18.69 & 46.01 & 60.59 & 34.09 \\ \midrule
MAGIC & 32.60 & 55.06 & 0.52 & 0.39 & 63.32 & 33.02 & 33.12 & 55.16 & 55.17 & 0.26 & 63.37 & 63.36 & 33.32 & 55.26 & 63.40 & 40.49 \\\midrule
M-SegFormer & 33.08 & 51.69 & 1.64 & 1.57 & 61.84 & 32.66 & 32.50 & 51.83  & 51.84 & 1.57 & 61.90 & 61.93 & 32.53 & 51.87 & \underline{61.92} & 39.36 \\  \midrule
AnySeg(SoTA) & 48.17 & 51.14 & 21.74 & 22.66 & 59.58 & 48.91 & 49.43 & 52.18 & 51.38 & 27.57 & 59.45 & 59.72 & 49.49 & 51.89 & \underline{59.41} & 47.13 \\ 
\midrule
Ours & \textbf{49.26} & \textbf{53.07} & \textbf{24.56} & \textbf{27.66} & \textbf{61.59} & \textbf{49.62} & \textbf{50.92} & \textbf{54.51} & \textbf{53.45} & \textbf{33.59} & \textbf{61.04} & \textbf{61.71} & \textbf{50.72} & \textbf{54.34} & \textbf{61.85} & \textbf{49.93} \\ 


\midrule
\textit{w.r.t} SoTA & +1.09 & +1.93 & +2.82 & +5.00 & +2.01 & +0.71 & +1.49 & +2.33 & +2.07 & +6.02 & +1.59 & +1.99 & +1.23 & +2.45 & +2.44 & +2.80 \\ 
\bottomrule
\end{tabular}}
\caption{Results of any-modality semantic segmentation based on the AnySeg's metric~\cite{zheng2024learning} are carried out on the DELIVER dataset using SegFormer-B0 as the backbone model.}
\vspace{-12pt}
\label{Tab:DELIVER}
\end{table*}

\noindent \textbf{Datasets.}
We evaluate on three multi-sensor datasets: \textbf{DELIVER}\cite{zhang2023DELIVERing} combines RGB, depth, LiDAR, event, and multi-view data across diverse weather (cloudy, foggy, night, rainy, sunny) and challenging conditions (motion blur, over/under-exposure, LiDAR jitter). It includes 25 semantic classes and serves as a key benchmark for robustness in autonomous driving. \textbf{MCubeS}\cite{Liang_2022_CVPR} offers 500 groups of multi-modal street scenes (RGB, AoLP, DoLP, NIR) with pixel-level material and semantic labels across 20 classes, split into 302 training, 96 validation, and 102 test groups at 1224×1024 resolution. \textbf{MUSES}~\cite{brodermann2024muses} targets adverse driving scenarios, providing 2,500 multi-sensor images (e.g., frame cameras, LiDAR, radar) under varying weather and lighting. 

\noindent \textbf{Backbone.} 
The framework used by \textbf{AnySeg}~\cite{zheng2024learning} is essentially a multi-modal \textbf{SegFormer}~\cite{xie2021segformer}, where multiple SegFormer encoders separately process information from different modalities before feeding into the segmentation head to make the final prediction. This modality-separated encoder structure is well-suited for self-distillation frameworks. AnySeg, based on this backbone, achieves state-of-the-art performance through a self-distillation approach.For convenience, we refer to it as \textbf{M-SegFormer} in the following discussion.
\textbf{CMNeXt}~\cite{zhang2023DELIVERing} uses dual branches to process the RGB modality and other modalities separately. It employs a Self-Query Hub to extract useful information from the additional modalities, and fuses features effectively through MHSA blocks and PPX blocks. Finally, the fused features are passed to the segmentation head to make the final prediction.

\noindent \textbf{Evaluation Metrics.}  
We follow AnySeg~\cite{zheng2024learning} for robustness evaluation with arbitrary missing modalities which can measure the independent representational capability of each modality. We also test on the sensor failure benchmark~\cite{liao2025benchmarking}, covering Entire-Missing Modality (EMM), Random-Missing Modality (RMM), and noisy modality (NM). They simulate three typical failure scenarios: complete sensor input loss, intermittent or partial sensor failure, and degraded or corrupted sensor data. Detailed metrics are provided in Appendix C. 

\noindent \textbf{Implementation}  
Experiments on DELIVER using M-SegFormer and CMNeXt as backbones are conducted on 2 NVIDIA L40 GPUs, while experiments on the MUSES and MCubeS datasets use 4 NVIDIA 3090 GPUs. We use AdamW with a learning rate of \(6\times10^{-5}\), a 10-epoch warm-up, and polynomial decay (exponent 0.9). The first training stage lasted for 200 epochs, and the second stage for 150 epochs. The batch size are 2 per GPU, with inputs cropped to 1024×1024. On the DELIVER dataset with the AnySeg backbone, the first stage takes 11 hours and the second stage takes 14 hours. Random seeds are fixed for reproducibility, with error margins within 0.1\% across different hardware.

\subsection{Results}
\label{sec:results}
\begin{figure*}[h!]
\centering
\includegraphics[width=1.0\textwidth]{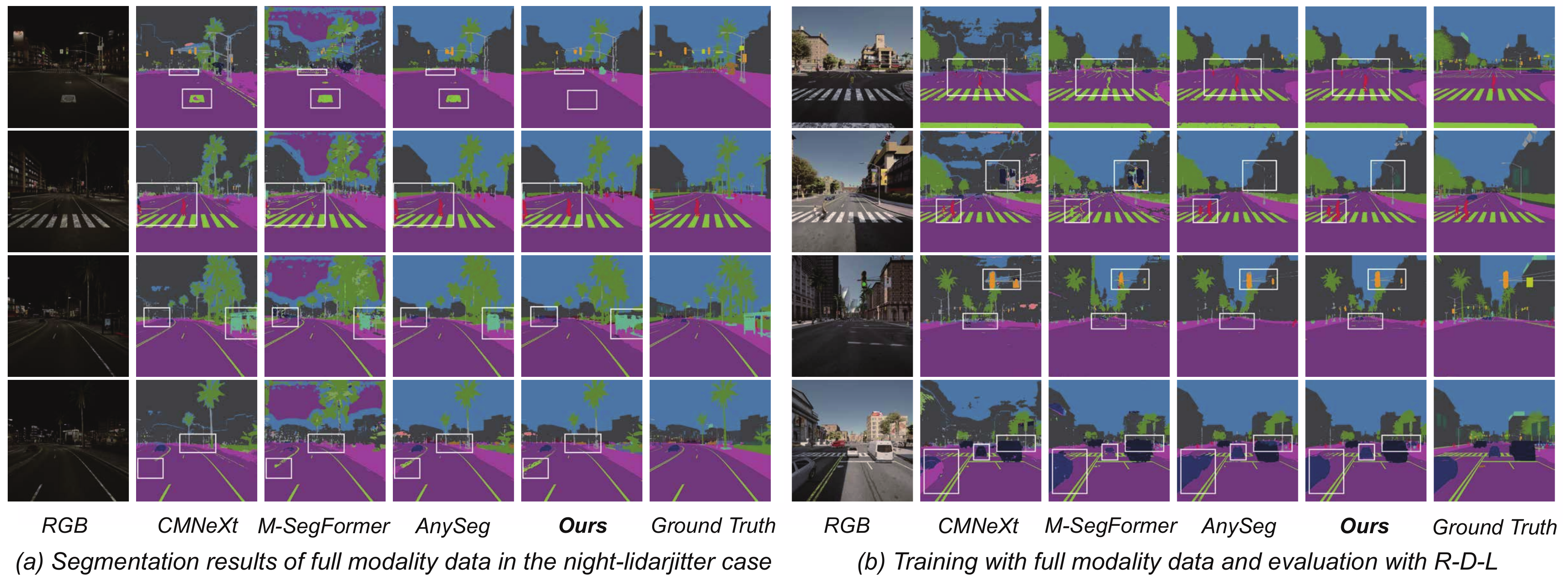}
\caption{Segmentation results of the model trained with our framework under full-modality and Event missing conditions.}
\label{fig:Result}
\end{figure*}

We built upon M-SegFormer baseline and compare it with several advanced methods for multi-modal semantic segmentation (MMSS), including M-Segformer, CMNeXt, MAGIC, MAGIC++, and AnySeg. We first evaluate the representational capability of arbitrary modality combinations on the DELIVER with the same evaluation metrics as AnySeg. The results are shown in Tab \ref{Tab:DELIVER}. 
As shown in Tab~\ref{Tab:DELIVER},
\name achieves highly competitive performance compared to SoTA methods, particularly in challenging scenarios with missing critical modalities. For example, when the Depth modality is absent, \name~significantly outperforms other models, achieving a +16.17\% mIoU improvement over MAGIC~\cite{zheng2024centering}. 
Though AnySeg represents a SoTA self-distillation framework in the MMSS domain, it suffers a performance drop of 2.51\% mIoU under full-modality input when built upon the M-SegFormer backbone. By contrast, \name~not only achieves superior robustness with a +2.80\% mIoU gain over AnySeg, but also maintains strong performance under full-modality conditions, with only a minor drop of 0.07\% mIoU. This demonstrates that our method significantly enhances model robustness while effectively preserving performance when all modalities are available.
We further conduct similar evaluations on the MCubeS and MUSES datasets, using the EMM and RMM metrics to simulate scenarios of complete and partial sensor input loss. As shown in Tab \ref{tab:datasets}, with a lightweight backbone (MiT-B0), our model outperforms most existing models that use the larger MiT-B2 backbone on these robustness metrics. 
Finally, we visualize the segmentation results under both full-modality and missing-modality conditions. As in Figure~\ref{fig:Result}, \name~demonstrates a fine-grained understanding of scene details in both cases.

\begin{table}[ht]
\centering
\renewcommand{\arraystretch}{1.2} 
\setlength{\tabcolsep}{1mm}       
\begin{tabular}{l|c|cc|cc|cc}
\midrule
\multirow{2}{*}{Model} & \multirow{2}{*}{MIT} & \multicolumn{2}{c|}{DELIVER} & \multicolumn{2}{c|}{MCubeS} & \multicolumn{2}{c}{MUSES} \\
\cmidrule{3-8}
 & & EMM & RMM & EMM & RMM & EMM & RMM \\
\midrule
CMNext   & B2 & 37.90 & 42.17 & 40.67 & 41.25 & 26.43 & 33.85 \\
Magic    & B2 & 44.97 & 45.30 & 42.61 & 42.25 & 33.34 & 42.49 \\
Magic++  & B2 & 44.85 & 47.06 & 42.53 & 44.61 & 33.17 & 43.24 \\
AnySeg   & B0 & 46.64 & 48.12 & 39.33 & 42.50 & 40.23 & 42.66 \\
Ours     & B0 & 49.93 & 51.90 & 43.22 & 44.18 & 41.12 & 42.78 \\
\midrule
\end{tabular}
\caption{Performance comparison of various advanced SegFormer-based models across different datasets, evaluated under EMM and RMM metrics.}
\label{tab:datasets}
\end{table}

\section{Ablation Studies}
\subsection{Analysis}
\begin{figure}[t]
\centering
\includegraphics[width=1\columnwidth]{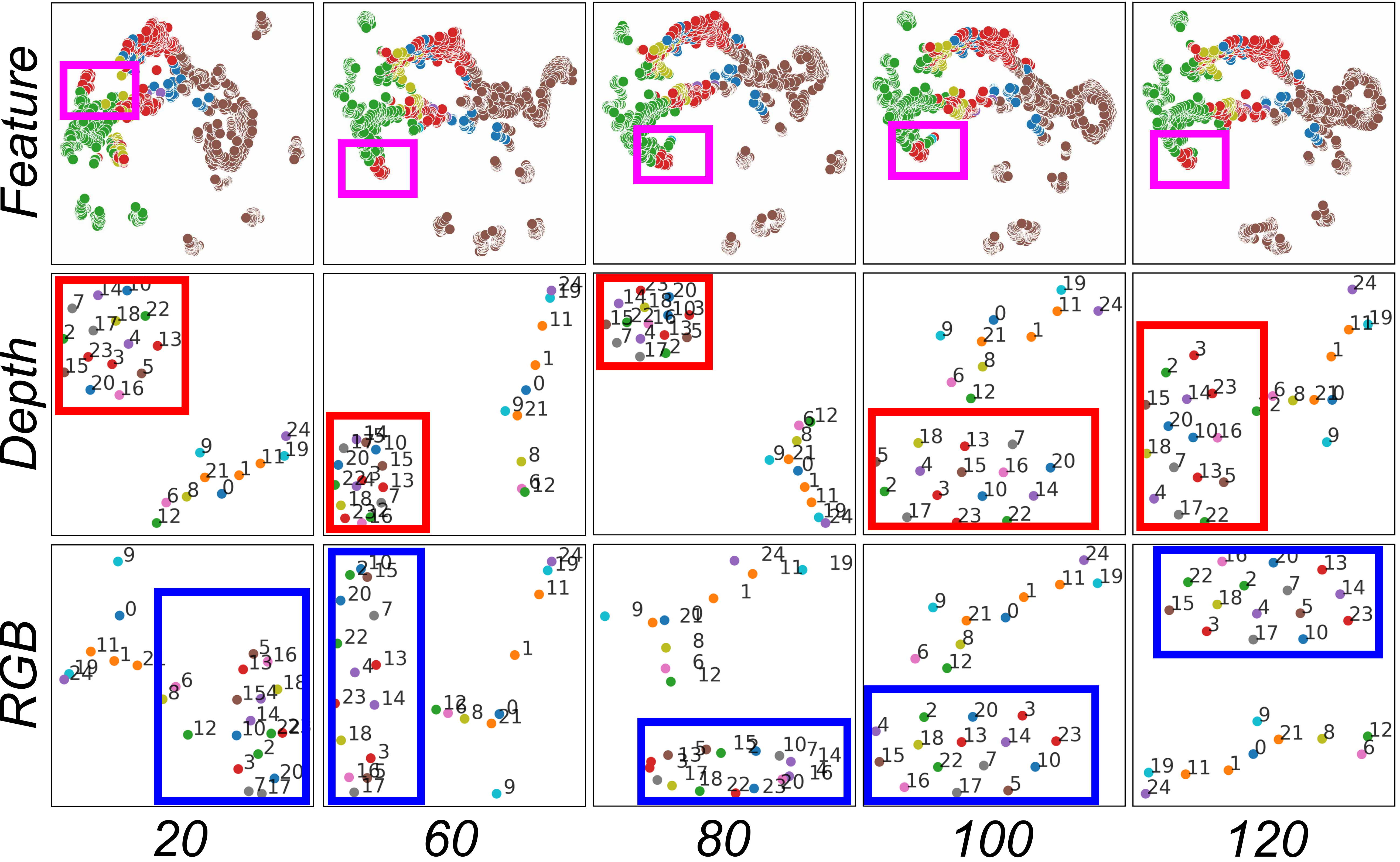}
\caption{This chart illustrates the visualization of model features across different numbers of epochs. The horizontal axis indicates the number of epochs, while the vertical axis (top to bottom) shows t-SNE visualizations of mixed features, Depth prototypes, and RGB prototypes.}
\label{fig:HPDM_ablaton}
\end{figure}
\noindent \textbf{Analysis of the Effectiveness of HPDM}
To further analyze the effectiveness of HPDM, we visualize the intermediate layer features of the model, as well as the prototype features of depth and RGB modalities separately. This aims to investigate the impact of HPDM on the interaction between the two modalities. The results are shown in Figure~\ref{fig:HPDM_ablaton}. First, by examining the feature maps in the first row, we observe that as training progresses, the features of different classes become increasingly separated. Meanwhile, as indicated by the purple box, the mixing between classes decreases over time. This demonstrates that the module has a positive impact on the final results. Next, we examine the prototype distributions in the second and third rows. Depth features tend to separate large-scale scene structures, while RGB features focus more on distinguishing fine details. Then, we analyze the influence of the Depth modality on the RGB modality. From step 20 to 80, as shown in the red box in the second row, the depth features gradually group the 25 classes into two broad categories — a characteristic of Depth prototypes. We also observe a similar trend in the third-row RGB modality, and the size of the blue boxes is positively correlated with the closeness to depth information. This indicates that the characteristics of depth prototypes are effectively transferred to the RGB modality.
Then, analyzing the influence of the RGB modality on the depth modality, during steps 80 to 120, the inter-class features of the Depth modality begin to diverge. This suggests that depth features have learned more fine-grained details from the RGB modality, leading to improved control over detail representation. This fusion of information makes the model more robust. For more details, please refer to Appendix D.

\noindent \textbf{Analysis of the Effectiveness of FSM}
To further analyze the effectiveness of FSM, we visualize a feature statistic that measures the proportion of feature patches originating from different teacher models. The results are shown in Figure~\ref{fig:result2}(b). We can observe that at the modality feature level, the FSM shows a relatively balanced reliance on the two teacher models. However, at the logits level, the FSM tends to favor the full-modality teacher model. This demonstrates that while the model maintains a balance between robustness and full-modality performance at the feature level, it prefers the full-modality features at the logits level, which are more beneficial for the final segmentation results.

\subsection{Ablation Study on the Two-Stage Framework}
To demonstrate the effectiveness and generality of our two-stage framework, we conduct experiments using M-SegFormer and CMNeXt models pretrained under full-modality settings. We evaluate them under entire-missing modality and full-modality conditions. Detailed results are shown in Tab~\ref{tab:2stage}. In terms of effectiveness, taking M-SegFormer as an example, we observe a significant improvement in the EMM metric after the first stage (+10.5\% mIoU), while its performance under full-modality conditions drops by 1.47\%. This validates the effectiveness of our first-stage design for enhancing robustness. In the second stage, under the guidance of the multi-teacher framework, the model maintains robustness comparable to that of the first stage, while its full-modality performance approaches that of the full-modality teacher model (with only a 0.07\% drop).

\noindent \textbf{Different Backbones.} To verify the generality of our framework, we conduct the same experiments with CMNeXt as backbone model. The results show that after the two-stage training, CMNeXt achieved a substantial improvement in robustness (+14.1\% mIoU), closely matching the robustness of the first-stage model (+0.06\% mIoU). Moreover, it maintains strong performance under full-modality conditions, with only a 0.23\% drop compared to the original full-modality model (\ie, \(60.12\xrightarrow{}59.89\) ).

\noindent \textbf{Training Efficiency.} The training cost depends on the backbone. For example, M-SegFormer requires 22GB GPU memory and 10 hours for standard training; in comparison, our first stage requires 31GB and 13 hours, and the second stage needs 40GB and 14 hours.

\begin{table}[ht]
\centering
\renewcommand{\arraystretch}{1.2} 
\setlength{\tabcolsep}{1mm}       
\begin{tabular}{l|cc|cc|cc}
\midrule
\multirow{2}{*}{Model} & \multicolumn{2}{c|}{Basic Model} & \multicolumn{2}{c|}{Stage1} & \multicolumn{2}{c}{Stage2} \\
\cmidrule{2-7}
& EMM & FULL & EMM & FULL & EMM & FULL \\
\midrule
M-SegFormer & 39.36 & 61.92 & 49.86 & 60.45 & 49.93 & 61.85 \\
CMNeXt      & 32.84 & 60.12 & 46.88 & 57.74 & 46.94 & 59.89 \\
\midrule
\end{tabular}
\caption{Performance comparison of models trained with the same M-SegFormer architecture at different stages. The basic model denotes the full-modality pretrained model.}
\label{tab:2stage}
\end{table}

\subsection{Ablation Study in Module Combinations}
To verify the effectiveness of HPDM in enhancing robustness and the ability of the FSM to preserve full-modality features, we construct a two-stage self-distillation baseline framework. In the first stage, we use the loss function defined in Eq. ~\ref{eq:lossorigin}, and in the second stage, we fix the loss weights of the full-modality teacher and the robustness teacher to 0.5 each. CMNeXt is selected as the base model for the experiments, Its full-modality performance, as shown in Table~\ref{tab:2stage}, is 60.12\%., the experimental results are shown in Table~\ref{tab:moduleab}. We then incorporate the HPDM and observe significant improvements in both EMM and RMM metrics (+4.00\% mIoU and + 1.51\% mIoU), indicating that HPDM effectively performs cross-modal distillation to enhance model robustness. Finally, we add the FSM, which leads to improvements in both full-modality and robustness metrics, demonstrating that FSM successfully transfers the strengths of multiple teacher models to the student model. 

\begin{table}[htbp]
\centering
\begin{tabular}{c|c|c|c|c|c|c}
\midrule
\textbf{Basic} & \textbf{HPDM} & \textbf{FSM} & \textbf{EMM} & \textbf{RMM} & \textbf{Full} \\ \midrule
\checkmark              &               &              & 43.45        & 51.69        & 57.50         \\ 
\midrule
\checkmark              & \checkmark             &              & 47.45        & 53.20        & 57.46         \\ \midrule
\checkmark              & \checkmark             & \checkmark            & 47.94        & 53.63        & 59.89         \\ \midrule
\end{tabular}
\caption{Performance results of the ablation study on the HPDM and FSM modules within the two-stage framework.}
\label{tab:moduleab}
\end{table}

\subsection{Ablation Study on Hyperparameters}
\begin{figure}[t!]
\centering
\includegraphics[width=1\columnwidth]{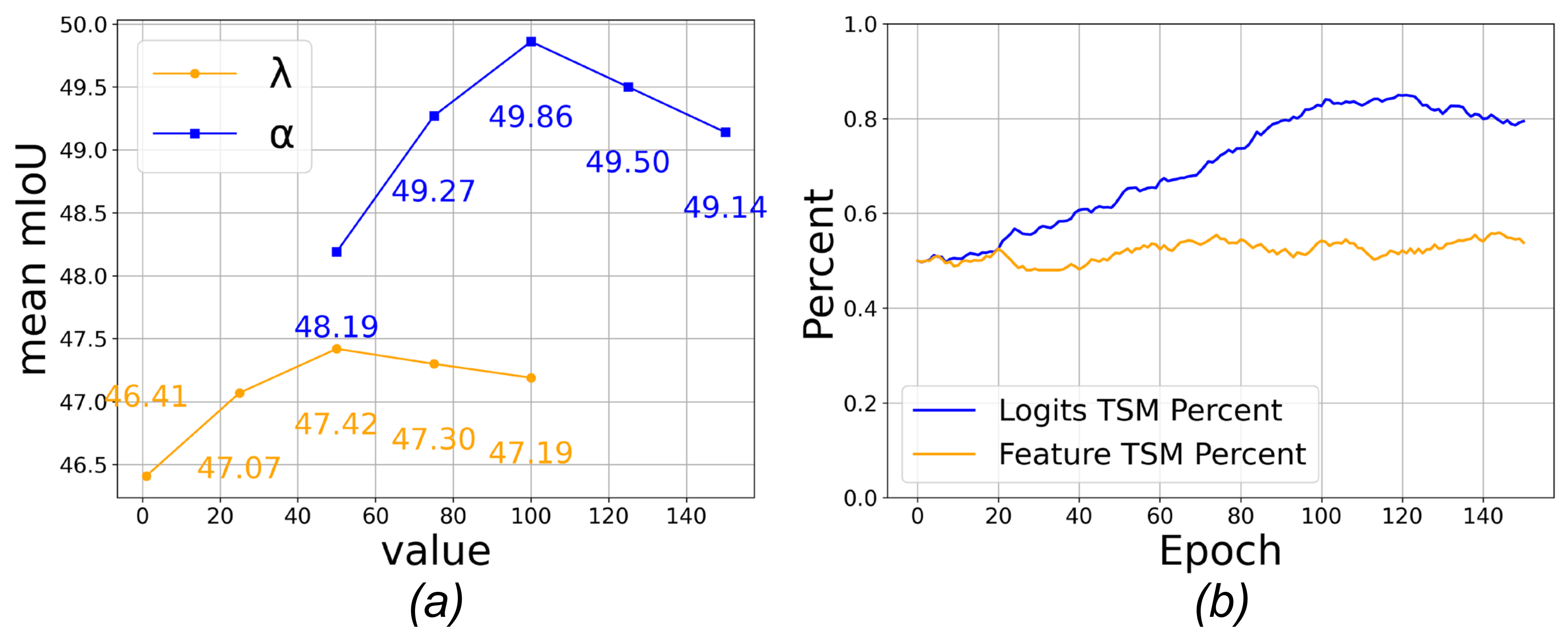}
\caption{Figure (a) illustrates the feature selection ratio from the full-modality teacher by the two Feature Selection Modules in the second stage. Figure (b) presents the ablation study under different values of the hyperparameters \(\lambda\) and \(\alpha\).}
\label{fig:result2}
\end{figure}

The hyperparameters we need to analyze are \(\lambda\) in Equation~\ref{eq:lossorigin} and \(\alpha\) in Equation~\ref{eq:losshp}, which respectively control the strength of modality-independent distillation and hybrid distillation. We use EMM as the evaluation metric and conduct robustness testing in the first stage. As shown in Figure~\ref{fig:result2}(a), we first test \(\lambda\) in the range of 0–100 without including HPDM (\ie, setting \(\alpha=0\)), and observe that the performance peaks at 47.42\% mIoU when \(\lambda=50\). We then fix it and perform an ablation study on \(\alpha\) within the range of 50–150. The results show that the performance reaches its peak at 49.86\% mIoU when \(\alpha=100\), which yields the optimal hyperparameter combination (\ie, \(\ \lambda =50,\alpha=100\)).

\section{Conclusion}
In conclusion, this paper proposes \name, a two-stage framework designed to enhance model robustness in a general and adaptable manner. Built upon a self-distillation and modality masking architecture, the framework incorporates Hybrid Prototype Distillation Module (HPDM) and Feature Selection Module (FSM). These components enable significant improvements in robustness while effectively mitigating performance degradation under full-modality conditions. Extensive experiments across two backbone networks and three datasets demonstrate the superiority of \name~over existing methods. This work marks a significant step forward in the study of robustness for multi-modal semantic segmentation (MMSS).

\bigskip
\noindent Thank you for reading these instructions carefully. We look forward to receiving your electronic files!

\bibliography{aaai2026}

\begin{thebibliography}{46}
\providecommand{\natexlab}[1]{#1}

\bibitem[{Alonso and Murillo(2019)}]{alonso2019ev}
Alonso, I.; and Murillo, A.~C. 2019.
\newblock EV-SegNet: Semantic segmentation for event-based cameras.
\newblock In \emph{Proceedings of the IEEE/CVF Conference on Computer Vision and Pattern Recognition Workshops}, 0--0.

\bibitem[{Ba and Caruana(2014)}]{ba2014deep}
Ba, J.; and Caruana, R. 2014.
\newblock Do deep nets really need to be deep?
\newblock \emph{Advances in neural information processing systems}, 27.

\bibitem[{Br{\"o}dermann et~al.(2024)Br{\"o}dermann, Bruggemann, Sakaridis, Ta, Liagouris, Corkill, and Van~Gool}]{brodermann2024muses}
Br{\"o}dermann, T.; Bruggemann, D.; Sakaridis, C.; Ta, K.; Liagouris, O.; Corkill, J.; and Van~Gool, L. 2024.
\newblock MUSES: The Multi-Sensor Semantic Perception Dataset for Driving under Uncertainty.
\newblock In \emph{European Conference on Computer Vision (ECCV)}.

\bibitem[{Camuffo, Mari, and Milani(2022)}]{camuffo2022recent}
Camuffo, E.; Mari, D.; and Milani, S. 2022.
\newblock Recent advancements in learning algorithms for point clouds: An updated overview.
\newblock \emph{Sensors}, 22(4): 1357.

\bibitem[{Chen et~al.(2014)Chen, Papandreou, Kokkinos, Murphy, and Yuille}]{chen2014semantic}
Chen, L.-C.; Papandreou, G.; Kokkinos, I.; Murphy, K.; and Yuille, A.~L. 2014.
\newblock Semantic image segmentation with deep convolutional nets and fully connected crfs.
\newblock \emph{arXiv preprint arXiv:1412.7062}.

\bibitem[{Chen et~al.(2018)Chen, Zhu, Papandreou, Schroff, and Adam}]{chen2018encoder}
Chen, L.-C.; Zhu, Y.; Papandreou, G.; Schroff, F.; and Adam, H. 2018.
\newblock Encoder-decoder with atrous separable convolution for semantic image segmentation.
\newblock In \emph{Proceedings of the European conference on computer vision (ECCV)}, 801--818.

\bibitem[{Chen et~al.(2021)Chen, Xian, Koepke, Shan, and Akata}]{chen2021distilling}
Chen, Y.; Xian, Y.; Koepke, A.; Shan, Y.; and Akata, Z. 2021.
\newblock Distilling audio-visual knowledge by compositional contrastive learning.
\newblock In \emph{Proceedings of the IEEE/CVF conference on computer vision and pattern recognition}, 7016--7025.

\bibitem[{Dancette et~al.(2021)Dancette, Cadene, Teney, and Cord}]{dancette2021beyond}
Dancette, C.; Cadene, R.; Teney, D.; and Cord, M. 2021.
\newblock Beyond question-based biases: Assessing multimodal shortcut learning in visual question answering.
\newblock In \emph{Proceedings of the IEEE/CVF International Conference on Computer Vision}, 1574--1583.

\bibitem[{Gao et~al.(2021)Gao, Pan, Li, Geng, and Zhao}]{gao2021we}
Gao, B.; Pan, Y.; Li, C.; Geng, S.; and Zhao, H. 2021.
\newblock Are we hungry for 3D LiDAR data for semantic segmentation? A survey of datasets and methods.
\newblock \emph{IEEE Transactions on Intelligent Transportation Systems}, 23(7): 6063--6081.

\bibitem[{Gat et~al.(2020)Gat, Schwartz, Schwing, and Hazan}]{gat2020removing}
Gat, I.; Schwartz, I.; Schwing, A.; and Hazan, T. 2020.
\newblock Removing bias in multi-modal classifiers: Regularization by maximizing functional entropies.
\newblock \emph{Advances in Neural Information Processing Systems}, 33: 3197--3208.

\bibitem[{Gu et~al.(2025)Gu, Yang, Feng, Wang, Zhang, Long, Chen, Cai, and Deng}]{gu2025breaking}
Gu, T.; Yang, K.; Feng, Z.; Wang, X.; Zhang, Y.; Long, D.; Chen, Y.; Cai, W.; and Deng, J. 2025.
\newblock Breaking the Modality Barrier: Universal Embedding Learning with Multimodal LLMs.
\newblock \emph{arXiv preprint arXiv:2504.17432}.

\bibitem[{Han et~al.(2021)Han, Wang, Su, Huang, and Tian}]{han2021greedy}
Han, X.; Wang, S.; Su, C.; Huang, Q.; and Tian, Q. 2021.
\newblock Greedy gradient ensemble for robust visual question answering.
\newblock In \emph{Proceedings of the IEEE/CVF international conference on computer vision}, 1584--1593.

\bibitem[{Hinton, Vinyals, and Dean(2015)}]{hinton2015distilling}
Hinton, G.; Vinyals, O.; and Dean, J. 2015.
\newblock Distilling the knowledge in a neural network.
\newblock \emph{arXiv preprint arXiv:1503.02531}.

\bibitem[{Hou et~al.(2024)Hou, Xing, Qian, Guo, Xin, Chen, Tang, Wang, Jiang, Liu et~al.}]{hou2024sdstrack}
Hou, X.; Xing, J.; Qian, Y.; Guo, Y.; Xin, S.; Chen, J.; Tang, K.; Wang, M.; Jiang, Z.; Liu, L.; et~al. 2024.
\newblock Sdstrack: Self-distillation symmetric adapter learning for multi-modal visual object tracking.
\newblock In \emph{Proceedings of the IEEE/CVF Conference on Computer Vision and Pattern Recognition}, 26551--26561.

\bibitem[{Huang et~al.(2022)Huang, Lin, Zhou, Yang, and Huang}]{huang2022modality}
Huang, Y.; Lin, J.; Zhou, C.; Yang, H.; and Huang, L. 2022.
\newblock Modality competition: What makes joint training of multi-modal network fail in deep learning?(provably).
\newblock In \emph{International conference on machine learning}, 9226--9259. PMLR.

\bibitem[{Kleinman, Achille, and Soatto(2023)}]{kleinman2023critical}
Kleinman, M.; Achille, A.; and Soatto, S. 2023.
\newblock Critical learning periods for multisensory integration in deep networks.
\newblock In \emph{Proceedings of the IEEE/CVF Conference on Computer Vision and Pattern Recognition}, 24296--24305.

\bibitem[{Li et~al.(2024)Li, Zhang, Zhao, Gao, and Li}]{li2024stitchfusion}
Li, B.; Zhang, D.; Zhao, Z.; Gao, J.; and Li, X. 2024.
\newblock Stitchfusion: Weaving any visual modalities to enhance multimodal semantic segmentation.
\newblock \emph{arXiv preprint arXiv:2408.01343}.

\bibitem[{Liang et~al.(2022)Liang, Wakaki, Nobuhara, and Nishino}]{Liang_2022_CVPR}
Liang, Y.; Wakaki, R.; Nobuhara, S.; and Nishino, K. 2022.
\newblock Multimodal Material Segmentation.
\newblock In \emph{Proceedings of the IEEE/CVF Conference on Computer Vision and Pattern Recognition (CVPR)}, 19800--19808.

\bibitem[{Liao et~al.(2025)Liao, Lei, Zheng, Moon, Wang, Wang, Paudel, Van~Gool, and Hu}]{liao2025benchmarking}
Liao, C.; Lei, K.; Zheng, X.; Moon, J.; Wang, Z.; Wang, Y.; Paudel, D.~P.; Van~Gool, L.; and Hu, X. 2025.
\newblock Benchmarking Multi-modal Semantic Segmentation under Sensor Failures: Missing and Noisy Modality Robustness.
\newblock \emph{arXiv preprint arXiv:2503.18445}.

\bibitem[{Lin et~al.(2017)Lin, Doll{\'a}r, Girshick, He, Hariharan, and Belongie}]{lin2017feature}
Lin, T.-Y.; Doll{\'a}r, P.; Girshick, R.; He, K.; Hariharan, B.; and Belongie, S. 2017.
\newblock Feature pyramid networks for object detection.
\newblock In \emph{Proceedings of the IEEE conference on computer vision and pattern recognition}, 2117--2125.

\bibitem[{Liu, Rabinovich, and Berg(2015)}]{liu2015parsenet}
Liu, W.; Rabinovich, A.; and Berg, A.~C. 2015.
\newblock Parsenet: Looking wider to see better.
\newblock \emph{arXiv preprint arXiv:1506.04579}.

\bibitem[{Liu, Guo, and S.~Lew(2016)}]{liu2016exploration}
Liu, Y.; Guo, Y.; and S.~Lew, M. 2016.
\newblock On the exploration of convolutional fusion networks for visual recognition.
\newblock In \emph{International conference on multimedia modeling}, 277--289. Springer.

\bibitem[{Long, Shelhamer, and Darrell(2015)}]{long2015fully}
Long, J.; Shelhamer, E.; and Darrell, T. 2015.
\newblock Fully convolutional networks for semantic segmentation.
\newblock In \emph{Proceedings of the IEEE conference on computer vision and pattern recognition}, 3431--3440.

\bibitem[{Noh, Hong, and Han(2015)}]{noh2015learning}
Noh, H.; Hong, S.; and Han, B. 2015.
\newblock Learning deconvolution network for semantic segmentation.
\newblock In \emph{Proceedings of the IEEE international conference on computer vision}, 1520--1528.

\bibitem[{Park et~al.(2019)Park, Kim, Lu, and Cho}]{park2019relational}
Park, W.; Kim, D.; Lu, Y.; and Cho, M. 2019.
\newblock Relational knowledge distillation.
\newblock In \emph{Proceedings of the IEEE/CVF conference on computer vision and pattern recognition}, 3967--3976.

\bibitem[{Passalis and Tefas(2018)}]{passalis2018learning}
Passalis, N.; and Tefas, A. 2018.
\newblock Learning deep representations with probabilistic knowledge transfer.
\newblock In \emph{Proceedings of the European conference on computer vision (ECCV)}, 268--284.

\bibitem[{Peng et~al.(2019)Peng, Jin, Liu, Li, Wu, Liu, Zhou, and Zhang}]{peng2019correlation}
Peng, B.; Jin, X.; Liu, J.; Li, D.; Wu, Y.; Liu, Y.; Zhou, S.; and Zhang, Z. 2019.
\newblock Correlation congruence for knowledge distillation.
\newblock In \emph{Proceedings of the IEEE/CVF international conference on computer vision}, 5007--5016.

\bibitem[{Peng et~al.(2022)Peng, Wei, Deng, Wang, and Hu}]{peng2022balanced}
Peng, X.; Wei, Y.; Deng, A.; Wang, D.; and Hu, D. 2022.
\newblock Balanced multimodal learning via on-the-fly gradient modulation.
\newblock In \emph{Proceedings of the IEEE/CVF conference on computer vision and pattern recognition}, 8238--8247.

\bibitem[{Reza, Prater-Bennette, and Asif(2024)}]{reza2024mmsformer}
Reza, M.~K.; Prater-Bennette, A.; and Asif, M.~S. 2024.
\newblock Mmsformer: Multimodal transformer for material and semantic segmentation.
\newblock \emph{IEEE Open Journal of Signal Processing}.

\bibitem[{Shin et~al.(2024)Shin, Lee, Kweon, and Oh}]{shin2024complementary}
Shin, U.; Lee, K.; Kweon, I.~S.; and Oh, J. 2024.
\newblock Complementary random masking for rgb-thermal semantic segmentation.
\newblock In \emph{2024 IEEE International Conference on Robotics and Automation (ICRA)}, 11110--11117. IEEE.

\bibitem[{Strudel et~al.(2021)Strudel, Garcia, Laptev, and Schmid}]{strudel2021segmenter}
Strudel, R.; Garcia, R.; Laptev, I.; and Schmid, C. 2021.
\newblock Segmenter: Transformer for semantic segmentation.
\newblock In \emph{Proceedings of the IEEE/CVF international conference on computer vision}, 7262--7272.

\bibitem[{Wang et~al.(2021)Wang, Wang, Li, and Wang}]{wang2021brief}
Wang, C.; Wang, C.; Li, W.; and Wang, H. 2021.
\newblock A brief survey on RGB-D semantic segmentation using deep learning.
\newblock \emph{Displays}, 70: 102080.

\bibitem[{Wang et~al.(2023)Wang, Yan, Zhang, Wei, Li, and Li}]{wang2023prototype}
Wang, S.; Yan, Z.; Zhang, D.; Wei, H.; Li, Z.; and Li, R. 2023.
\newblock Prototype knowledge distillation for medical segmentation with missing modality.
\newblock In \emph{ICASSP 2023-2023 IEEE International Conference on Acoustics, Speech and Signal Processing (ICASSP)}, 1--5. IEEE.

\bibitem[{Wang et~al.(2020)Wang, Zhou, Jiang, Bai, and Xu}]{wang2020intra}
Wang, Y.; Zhou, W.; Jiang, T.; Bai, X.; and Xu, Y. 2020.
\newblock Intra-class feature variation distillation for semantic segmentation.
\newblock In \emph{Computer Vision--ECCV 2020: 16th European Conference, Glasgow, UK, August 23--28, 2020, Proceedings, Part VII 16}, 346--362. Springer.

\bibitem[{Xia et~al.(2023)Xia, Li, Deng, Xiong, Dou, and Hu}]{xia2023robust}
Xia, W.; Li, X.; Deng, A.; Xiong, H.; Dou, D.; and Hu, D. 2023.
\newblock Robust cross-modal knowledge distillation for unconstrained videos.
\newblock \emph{arXiv preprint arXiv:2304.07775}.

\bibitem[{Xie et~al.(2021)Xie, Wang, Yu, Anandkumar, Alvarez, and Luo}]{xie2021segformer}
Xie, E.; Wang, W.; Yu, Z.; Anandkumar, A.; Alvarez, J.~M.; and Luo, P. 2021.
\newblock SegFormer: Simple and efficient design for semantic segmentation with transformers.
\newblock \emph{Advances in neural information processing systems}, 34: 12077--12090.

\bibitem[{Zhang et~al.(2023{\natexlab{a}})Zhang, Liu, Yang, Hu, Liu, and Stiefelhagen}]{zhang2023cmx}
Zhang, J.; Liu, H.; Yang, K.; Hu, X.; Liu, R.; and Stiefelhagen, R. 2023{\natexlab{a}}.
\newblock CMX: Cross-modal fusion for RGB-X semantic segmentation with transformers.
\newblock \emph{IEEE Transactions on intelligent transportation systems}.

\bibitem[{Zhang et~al.(2023{\natexlab{b}})Zhang, Liu, Shi, Yang, Rei{\ss}, Peng, Fu, Wang, and Stiefelhagen}]{zhang2023DELIVERing}
Zhang, J.; Liu, R.; Shi, H.; Yang, K.; Rei{\ss}, S.; Peng, K.; Fu, H.; Wang, K.; and Stiefelhagen, R. 2023{\natexlab{b}}.
\newblock Delivering arbitrary-modal semantic segmentation.
\newblock In \emph{Proceedings of the IEEE/CVF Conference on Computer Vision and Pattern Recognition}, 1136--1147.

\bibitem[{Zhao, Wang, and Wei(2024)}]{zhao2024mitigating}
Zhao, S.; Wang, X.; and Wei, X. 2024.
\newblock Mitigating accuracy-robustness trade-off via balanced multi-teacher adversarial distillation.
\newblock \emph{IEEE transactions on pattern analysis and machine intelligence}, 46(12): 9338--9352.

\bibitem[{Zheng et~al.(2021)Zheng, Lu, Zhao, Zhu, Luo, Wang, Fu, Feng, Xiang, Torr et~al.}]{zheng2021rethinking}
Zheng, S.; Lu, J.; Zhao, H.; Zhu, X.; Luo, Z.; Wang, Y.; Fu, Y.; Feng, J.; Xiang, T.; Torr, P.~H.; et~al. 2021.
\newblock Rethinking semantic segmentation from a sequence-to-sequence perspective with transformers.
\newblock In \emph{Proceedings of the IEEE/CVF conference on computer vision and pattern recognition}, 6881--6890.

\bibitem[{Zheng et~al.(2024{\natexlab{a}})Zheng, Lyu, Jiang, Zhou, Wang, and Hu}]{zheng2024magic++}
Zheng, X.; Lyu, Y.; Jiang, L.; Zhou, J.; Wang, L.; and Hu, X. 2024{\natexlab{a}}.
\newblock MAGIC++: Efficient and Resilient Modality-Agnostic Semantic Segmentation via Hierarchical Modality Selection.
\newblock \emph{arXiv preprint arXiv:2412.16876}.

\bibitem[{Zheng et~al.(2024{\natexlab{b}})Zheng, Lyu, Zhou, and Wang}]{zheng2024centering}
Zheng, X.; Lyu, Y.; Zhou, J.; and Wang, L. 2024{\natexlab{b}}.
\newblock Centering the value of every modality: Towards efficient and resilient modality-agnostic semantic segmentation.
\newblock In \emph{European Conference on Computer Vision}, 192--212. Springer.

\bibitem[{Zheng et~al.(2024{\natexlab{c}})Zheng, Xue, Chen, Yan, Jiang, Lyu, Yang, Zhang, and Hu}]{zheng2024learning}
Zheng, X.; Xue, H.; Chen, J.; Yan, Y.; Jiang, L.; Lyu, Y.; Yang, K.; Zhang, L.; and Hu, X. 2024{\natexlab{c}}.
\newblock Learning Robust Anymodal Segmentor with Unimodal and Cross-modal Distillation.
\newblock \emph{arXiv preprint arXiv:2411.17141}.

\bibitem[{Zhou et~al.(2020)Zhou, Qi, Wan, Huang, and Yang}]{zhou2020rgb}
Zhou, H.; Qi, L.; Wan, Z.; Huang, H.; and Yang, X. 2020.
\newblock RGB-D co-attention network for semantic segmentation.
\newblock In \emph{Proceedings of the Asian conference on computer vision}.

\bibitem[{Zhou et~al.(2023)Zhou, Liu, Hu, Zhou, and Ma}]{zhou2023unidistill}
Zhou, S.; Liu, W.; Hu, C.; Zhou, S.; and Ma, C. 2023.
\newblock Unidistill: A universal cross-modality knowledge distillation framework for 3D object detection in bird's-eye view.
\newblock In \emph{Proceedings of the IEEE/CVF conference on computer vision and pattern recognition}, 5116--5125.

\bibitem[{Zhou et~al.(2022)Zhou, Wang, Konukoglu, and Van~Gool}]{zhou2022rethinking}
Zhou, T.; Wang, W.; Konukoglu, E.; and Van~Gool, L. 2022.
\newblock Rethinking semantic segmentation: A prototype view.
\newblock In \emph{Proceedings of the IEEE/CVF conference on computer vision and pattern recognition}, 2582--2593.

\end{thebibliography}

\end{document}